\documentclass[11pt,a4paper]{article}
\usepackage[hyperref]{acl2019}
\usepackage{times}
\usepackage{latexsym}
\usepackage{amsmath}
\usepackage{multirow}
\usepackage{multicol}
\usepackage{array}
\usepackage{booktabs}
\usepackage{url}
\usepackage{color, soul}

\usepackage{hhline}
\usepackage[T1]{fontenc}
\usepackage{graphicx}
\usepackage{eurosym}
\usepackage{adjustbox}
\usepackage{kantlipsum}
\usepackage{float}
\usepackage{tabularx}
\graphicspath{ {images/} }
\usepackage{listings}
\usepackage{stfloats}
\usepackage{csquotes}
\usepackage[super]{nth}
\usepackage[inline,shortlabels]{enumitem}
\usepackage{subfig}
\usepackage{enumitem}

\usepackage{relsize}
\usepackage{hyperref}

\usepackage{svg}
\usepackage{makecell}
\usepackage{enumitem}
\usepackage[scientific-notation=true]{siunitx}
\usepackage{marginnote}
\usepackage{amssymb}
\usepackage{pifont}
\usepackage[textsize=tiny]{todonotes}
\usepackage{mdframed}
\usepackage{dirtytalk}
\usepackage{xprintlen}

\usepackage{todonotes}

\aclfinalcopy 
\def\aclpaperid{1776} 

\title{Attention-based Conditioning Methods \\
for External Knowledge Integration}

\author{ 
	Katerina Margatina$^{1}$, 
	Christos Baziotis$^{2}$
	\thanks{\,\, The research was conducted when the author was a researcher at School of ECE, NTUA in Athens, Greece.}, 
	Alexandros Potamianos$^{1,3,4}$ \\\\
	$^1$School of ECE, National Technical University of Athens, Athens, Greece \\
	$^2$ School of Informatics, University of Edinburgh, UK \\
	$^3$ Signal Analysis and Interpretation Laboratory (SAIL), USC, Los Angeles, USA\\
	$^4$ Behavioral Signal Technologies, Los Angeles, USA\\
	{\tt  el12108@central.ntua.gr, c.baziotis@sms.ed.ac.uk,} \\
	{\tt potam@central.ntua.gr}
	}

\date{2019}

\begin{document}

\maketitle
\begin{abstract}
    In this paper, we present a novel approach for incorporating external knowledge in Recurrent Neural Networks (RNNs).
    We propose the integration of lexicon features into the self-attention mechanism of RNN-based architectures.
    This form of conditioning on the attention distribution, enforces the contribution of the most salient words for the task at hand.
    We introduce three methods, namely attentional concatenation, feature-based gating and affine transformation. 
    Experiments on six benchmark datasets show the effectiveness of our methods. 
    Attentional feature-based gating yields consistent performance improvement across tasks.
    Our approach is implemented as a simple add-on module for RNN-based models with minimal computational overhead and can be adapted to any deep neural architecture.

\end{abstract}

\section{Introduction}

Modern deep learning algorithms often do away with feature engineering and learn latent representations directly from raw data that are given as input
to Deep Neural Networks (DNNs)~\cite{Mikolov:2013:DRW:2999792.2999959, NIPS2017_7209, N18-1202}.
However, it has been shown that linguistic knowledge (manually or semi-automatically encoded into lexicons and knowledge bases) can significantly improve DNN performance for Natural Language Processing (NLP) tasks, 
such as natural language inference~\cite{Q17-1022}, 
language modelling~\cite{ahn2016neural}, 
named entity recognition~\cite{C18-1161} and relation extraction~\cite{D18-1157}.

For NLP tasks, external sources of information are typically incorporated into deep neural architectures by processing the raw input \textit{in the context} of such external linguistic knowledge.
In machine learning, this contextual processing is known as \textit{conditioning}; the computation carried out by a model is conditioned or modulated by information extracted from an auxiliary input. 
The most commonly-used method of conditioning is concatenating a representation of the external information to the input or hidden network layers.

Attention mechanisms~\cite{bahdanau2014neural,NIPS2017_7181,lin+al-2017-embed-iclr} are a key ingredient for achieving state-of-the-art performance in tasks such as textual entailment ~\cite{DBLP:journals/corr/RocktaschelGHKB15}, question answering \cite{wei2018fast}, and neural machine translation~\cite{DBLP:journals/corr/WuSCLNMKCGMKSJL16}. 
Often task-specific attentional architectures are proposed in the literature to further improve DNN performance
~\cite{P17-1168, pmlr-v37-xuc15, K18-1030}. 

In this work, we propose a novel way of utilizing word-level prior information encoded in linguistic, sentiment, and emotion lexicons, to improve classification performance.
Usually, lexicon features are concatenated to word-level representations~\cite{Wang:2016:EER:3060832.3061031, D17-1211,DBLP:journals/corr/abs-1804-07000}, as additional features to the embedding of each word or the hidden states of the model.
By contrast, we propose to incorporate them into the self-attention mechanism of RNNs.
Our goal is to enable the self-attention mechanism to identify the most informative words, by directly conditioning on their additional lexicon features.

Our contributions are the following:
(1) we propose an alternative way for incorporating external knowledge to RNN-based architectures,
(2) we present empirical results that our proposed approach consistently outperforms strong baselines, and 
(3) we report state-of-the-art performance in two datasets.
We make our source code publicly available\footnote{\url{https://github.com/mourga/affective-attention}}.

\section{Related Work}
In the traditional machine learning literature where statistical models are based on sparse features, affective lexicons
have been shown to be highly effective for tasks such as sentiment analysis, as they provide additional information not captured in the raw training data \cite{DBLP:conf/kdd/HuL04,DBLP:conf/coling/KimH04,Ding:2008:HLA:1341531.1341561, P14-2089, Taboada:2011:LMS:2000517.2000518}. 
After the emergence of pretrained word representations~\cite{Mikolov:2013:DRW:2999792.2999959, pennington2014glove}, the use of lexicons is no longer common practice, since word embeddings can also capture some of the affective meaning of these words. 

Recently, there have been notable contributions towards integrating linguistic knowledge into DNNs for various NLP tasks. 
For sentiment analysis, \citet{teng-etal-2016-context} integrate lexicon features to an RNN-based model with a custom weighted-sum calculation of word features.
\citet{W17-5220} propose three convolutional neural network specific methods of lexicon integration
achieving state-of-the-art performance on two datasets.
\citet{N18-2041} concatenate features from a knowledge base to word representations in an attentive bidirectional LSTM architecture, also reporting state-of-the-art results.
For sarcasm detection, \citet{D17-1211} incorporate psycholinguistic, stylistic, structural, and readability features by concatenating them to paragraph and document-level representations.

Furthermore, there is limited literature regarding the development and evaluation of methods for combining representations in deep neural networks.
\citet{DBLP:journals/corr/PetersABP17} claim that concatenation, non-linear mapping and attention-like mechanisms are unexplored methods for including language model representations in their sequence model. 
They employ simple concatenation, leaving the exploration of other methods to future work.
\citet{dumoulin2018feature-wise} provide an overview of feature-wise transformations such as concatenation-based conditioning, conditional biasing and gating mechanisms. 
They review the effectiveness of conditioning methods in tasks such as visual question answering~\cite{Strub_2018_ECCV}, style transfer~\cite{45832} and language modeling~\cite{Dauphin:2017:LMG:3305381.3305478}.
They also extend the work by~\citet{DBLP:journals/corr/abs-1709-07871}, which proposes the Feature-wise Linear Modulation (FiLM) framework, and investigate its applications in visual reasoning tasks. 
\citet{balazs-matsuo-2019-gating} provide an empirical study showing the effects of different ways of combining character and word representations in word-level and sentence-level evaluation tasks.
Some of the reported findings are that gating conditioning performs consistently better across a variety of word similarity and relatedness tasks.

\section{Proposed Model}\label{sec:model}

\subsection{Network Architecture}
\label{sec:network}

\noindent\textbf{Word Embedding Layer}. The input sequence of words $w_1,w_2,...,w_T$ is projected to a low-dimensional vector space $ R^W$, where $W$ is the size of the embedding layer and $T$ the number of words in a sentence. 
We initialize the weights of the embedding layer with pretrained word embeddings.

\vspace{3pt}
\noindent\textbf{LSTM Layer}. A Long Short-Term Memory unit (LSTM) \cite{hochreiter1997lstm} takes as input the words of a sentence and produces the word annotations $h_1,h_2,...,h_T$, where $h_i$ is the hidden state of the LSTM at time-step $i$, summarizing  all sentence information up to $w_i$. 

\vspace{3pt}
\noindent\textbf{Self-Attention Layer}.
We use a self-attention mechanism~\cite{D16-1053} to find the relative importance of each word for the task at hand.
The attention mechanism assigns a score $a_i$ to each word annotation $h_i$. We compute the fixed representation $r$ of the input sequence, as the weighted sum of all the word annotations. Formally:
\begin{align}
a_i &= softmax(v_a^\intercal f(h_i))\label{eq:ai} \\
r &= \sum_{i=1}^{T} a_i h_i \label{eq:r}   
\end{align}
where $f(.)$ corresponds to a non-linear transformation $tanh(W_a h_i + b_a)$ and $W_a, b_a, v_a$ are the parameters of the attention layer. 

\begin{table}[tbh]
\centering
\resizebox{\columnwidth}{!}{%
\begin{tabular}{cccc}
\Xhline{2\arrayrulewidth}
Lexicons & Annotations  & \# dim. & \# words\\ \hhline{====}
LIWC  & psycho-linguistic & 73 & 18,504  \\ \hline
Bing Liu   & valence & 1  & 2,477 \\ \hline
AFINN  & sentiment & 1  & 6,786 \\ \hline
MPQA  & sentiment & 4 & 6,886 \\ \hline
SemEval15  & sentiment & 1 & 1,515  \\ \hline
Emolex & emotion & 19 & 14,182 \\\Xhline{2\arrayrulewidth}
\end{tabular}
}
\caption{The lexicons used as external knowledge.}
\label{table:lexicons}
\end{table}
\subsection{External Knowledge}
In this work, we augment our models with existing linguistic and affective knowledge from human experts.
Specifically, we leverage lexica containing psycho-linguistic, sentiment and emotion annotations. 
We construct a feature vector $c(w_i)$ for every word in the vocabulary by concatenating the word's annotations from the lexicons shown in Table~\ref{table:lexicons}. 
For missing words we append zero in the corresponding dimension(s) of $c(w_i)$.

\subsection{Conditional Attention Mechanism}
\label{sec:cond_attention}
\begin{figure}[h!]%
	\captionsetup{farskip=0pt} 
    \subfloat[Attentional Concatenation]{{\includegraphics[width=\columnwidth]{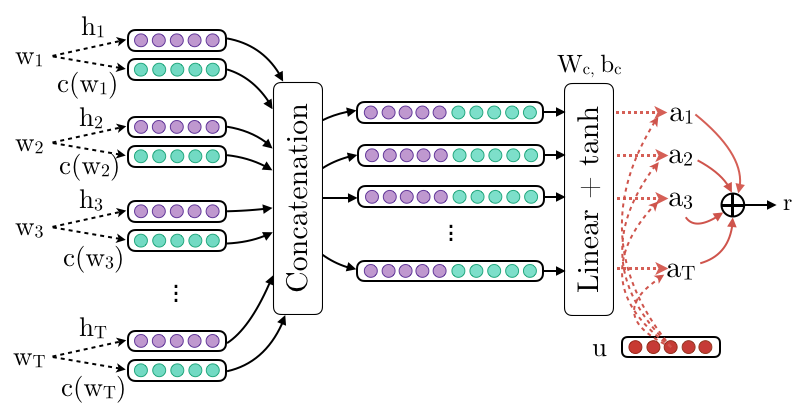} }}%
    \label{fig:conc}
    \subfloat[Attentional Feature-Based Gating]{{\includegraphics[width=\columnwidth]{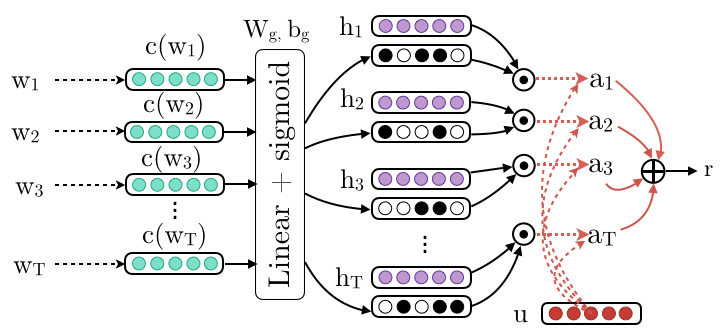} }}%
    \label{fig:gating_big}
    \subfloat[Attentional Affine Transformation]{{\includegraphics[width=\columnwidth]{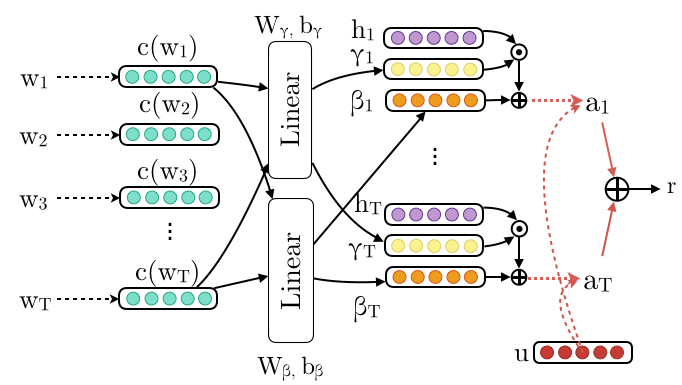} }}%
    \label{fig:affine}
    \caption{The proposed conditioning methods of the self-attention mechanism.}%
    \label{fig:cond_methods}%
\end{figure}

We extend the standard self-attention mechanism (Eq.~\ref{eq:ai},~\ref{eq:r}), in order to condition the attention distribution of a given sentence, on each word's prior lexical information.
To this end, we use as input to the self-attention layer both the word annotation $h_i$, as well as the lexicon feature $c(w_i)$ of each word.
Therefore, we replace $f(h_i)$ in Eq.~\ref{eq:ai} with $f(h_i, c(w_i))$.
Specifically, we explore three conditioning methods, which are illustrated in Figure~\ref{fig:cond_methods}. 
We refer to the conditioning function as $f_i(.)$, the weight matrix as $W_i$ and the biases as $b_i$, where $i$ is an indicative letter for each method.
We present our results in Section~\ref{sec:results} (Table~\ref{table:results}) and we denote the three conditioning methods as \textit{``conc.''}, \textit{``gate''} and \textit{``affine''} respectively.

 \vspace{3pt}
\noindent\textbf{Attentional Concatenation.} 
In this approach, as illustrated in Fig.~\ref{fig:cond_methods}(a), we learn a function of the concatenation of each word annotation $h_i$ with its lexicon features $c(w_i)$. 
The intuition is that by adding extra dimensions to $h_i$, learned representations are more discriminative. 
Concretely:
\begin{align}
f_c(h_i, c(w_i)) &= tanh(W_c [h_i \parallel c(w_i)] + b_c)\label{eq:f_concat}
\end{align}
where $\parallel$ denotes the concatenation operation and $W_c, b_c$ are learnable parameters.

\vspace{3pt}
\noindent\textbf{Attentional Feature-based Gating.} 
The second approach, illustrated in Fig.~\ref{fig:cond_methods}(b), learns a feature mask, which is applied on each word annotation $h_i$. 
Specifically, a gate mechanism with a sigmoid activation function, generates a mask-vector from each $c(w_i)$ with values between 0 and 1 (black and white dots in Fig.~\ref{fig:cond_methods}(b)). 
Intuitively, this gating mechanism selects salient dimensions (i.e. features) of $h_i$, conditioned on the lexical information. Formally:
\begin{align}
f_g(h_i, c(w_i)) &= \sigma(W_g c(w_i) + b_g) \odot h_i \label{eq:f_gate}
\end{align}
where $\odot$ denotes element-wise multiplication and $W_g, b_g$ are learnable parameters.

\vspace{3pt}
\noindent\textbf{Attentional Affine Transformation.} The third approach, shown in Fig.~\ref{fig:cond_methods}(c), is adopted from the work of~\citet{DBLP:journals/corr/abs-1709-07871} and applies a feature-wise affine transformation to the latent space of the hidden states. 
Specifically, we use the lexicon features $c(w_i)$, in order to conditionally generate the corresponding scaling $\gamma(\cdot)$ and shifting $\beta(\cdot)$ vectors.
Concretely:
\begin{gather}
f_a(h_i, c(w_i)) = \gamma(c(w_i)) \odot h_i + \beta(c(w_i))\label{eq:f_affine} \\
\gamma(x) = W_\gamma x + b_\gamma, \quad \beta(x) = W_\beta x + b_\beta \label{eq:gamma}
\end{gather}
where $W_\gamma, W_\beta, b_\gamma, b_\beta$ are learnable parameters.

\subsection{Baselines}\label{sec:baselines}

We employ two baselines:
The first baseline is an LSTM-based architecture 
augmented with a self-attention mechanism (Sec.~\ref{sec:network})
with no external knowledge.
The second baseline 
incorporates lexicon information by
concatenating the $c(w_i)$ vectors to the word representations in the embedding layer.
In Table~\ref{table:results} we use the abbreviations \textit{``baseline''} and \textit{``emb. conc.''} for the two baseline models respectively.

\section{Experiments}
\noindent\textbf{Lexicon Features}. As prior knowledge, we leverage the lexicons presented in Table~\ref{table:lexicons}. We selected widely-used lexicons that represent different facets of affective and psycho-linguistic features, namely;  LIWC~\cite{doi:10.1177/0261927X09351676}, Bing Liu Opinion Lexicon~\cite{DBLP:conf/kdd/HuL04}, AFINN~\cite{IMM2011-06010}, Subjectivity Lexicon~\cite{Wilson:2005:OSS:1225733.1225751}, 
SemEval 2015 English Twitter Lexicon~\cite{NRCJAIR14}, and NRC Emotion Lexicon (EmoLex)~\cite{Mohammad13}.

\vspace{3pt}
\noindent\textbf{Datasets}. The proposed framework can be applied to different domains and tasks. 
In this paper, we experiment with sentiment analysis, emotion recognition, irony, and sarcasm detection. 
Details of the benchmark datasets are shown in Table~\ref{table:datasets}.

\begin{table*}[t!]
\centering
\footnotesize
{\renewcommand{\arraystretch}{1.4}
\begin{tabularx}{\textwidth}{lXllrrr}
\Xhline{2\arrayrulewidth}
Dataset & Study & Task      & Domain        & Classes & \textit{$N_{train}$} & $N_{test}$  \\ \hline\hline

SST-5      &  ~\citet{socher-EtAl:2013:EMNLP}        & Sentiment   & Movie Reviews & 5 & 9,645  & 2,210  \\
Sent17     &  ~\citet{S17-2088}                      & Sentiment   & Twitter       & 3 & 49,570 & 12,284 \\ \hline
PhychExp   & ~\citet{doi:10.1177/053901886025004001} & Emotion     & Experiences   & 7 & 1000   & 6480\\ \hline
Irony18    &  ~\citet{S18-1005}                      & Irony       & Twitter       & 4 & 3,834  & 784    \\ \hline
SCv1       & ~\citet{lukin-walker:2013:LASM}         & Sarcasm     & Debate Forums & 2 & 1000   & 995    \\
SCv2       &   ~\citet{W16-3604}                     & Sarcasm     & Debate Forums & 2 & 1000   & 2260  \\ \Xhline{2\arrayrulewidth}

\end{tabularx}
}
\caption{Description of benchmark datasets. We split 10\% of the train set to serve as the validation set.}
\label{table:datasets}
\end{table*}

\begin{table*}[t!]
\small
\centering
\begin{adjustbox}{width=\textwidth}
{\renewcommand{\arraystretch}{1.4}
\setlength{\tabcolsep}{2pt}
\begin{tabular}{r|cccccc}
\Xhline{2\arrayrulewidth}
Model        & SST-5 & Sent17  & PhychExp  & Irony18  & SCv1   & SCv2  \\ \hline \hline
baseline     & $43.5\pm0.5$  & $68.3\pm0.2$   & $53.2\pm0.8$      & $46.3\pm1.4$     & $64.1\pm0.5$   & $74.0\pm0.7$  \\
emb. conc.     & $43.3\pm0.6$  & $68.4\pm0.2$    & $57.1\pm1.2$      & $48.1\pm1.2$     & $64.2\pm0.7$   & $74.2\pm0.7$   \\ \hline
conc.      & $44.0\pm0.7$  & $68.6\pm0.3$    & $54.3\pm0.6$      & $47.4\pm0.9$          & $\textbf{65.1}\pm0.6$    & $74.3\pm1.2$  \\
gate       & $44.2\pm0.4$  & $68.7\pm0.3$    & $53.4\pm1.0$      & $\textbf{48.5}\pm0.7$ & $64.7\pm0.7$   & $74.3\pm1.2$  \\
affine       & $43.2\pm0.7$  & $68.5\pm0.3$    & $53.1\pm0.9$      & $45.3\pm1.5$          & $60.3\pm0.8$   & $74.0\pm1.0$ \\ \hline
gate+emb.conc. & $\textbf{46.2}\pm0.5$ & $\textbf{68.9}\pm0.3$ & $\textbf{57.2}\pm1.1$ & $\textbf{48.4}\pm1.0$          & $\textbf{64.9}\pm0.6$ & $\textbf{74.4}\pm0.9$ \\ \hline
\multirow{2}{*}{state-of-the-art}     & $51.7$          & $68.5$          & $57.0$          & $53.6$          & $69.0$          & $76.0$          \\
      & \citet{Shen2018DiSANDS} & \citet{S17-2094} & \citet{D17-1169} & \citet{S18-1100} & \citet{D17-1169} & \citet{W18-6202} \\
\Xhline{2\arrayrulewidth}
\end{tabular}
}
\end{adjustbox}
\caption{Comparison across benchmark datasets. Reported values are averaged across ten runs. All reported measures are $F_1$ scores, apart from $SST-5$ which is evaluated with \textit{Accuracy}. }
\label{table:results}
\end{table*}

\vspace{3pt}
\noindent\textbf{Preprocessing}.  
To preprocess the words, we use the tool $Ekphrasis$~\citep{S17-2126}. After tokenization, we map each word to the corresponding pretrained word representation: Twitter-specific word2vec embeddings~\cite{W18-6209} for the Twitter datasets, and fasttext~\cite{bojanowski2017enriching} for the rest.

\vspace{3pt}
\noindent\textbf{Experimental Setup}. 
For all methods, we employ a single-layer LSTM model with 300 neurons augmented with a self-attention mechanism, as described in Section~\ref{sec:model}. 
As regularization techniques, we apply early stopping, Gaussian noise $N(0, 0.1)$ to the word embedding layer, 
and dropout to the LSTM layer with $p=0.2$.
We use Adam to optimize our networks~\cite{kingma2014} 
with mini-batches of size 64 
and clip the norm of the gradients~\cite{pascanu2013a} at 0.5, as an extra safety measure against exploding gradients. 
We also use PyTorch \cite{paszke2017automatic} and scikit-learn \cite{pedregosa2011}.

\section{Results \& Analysis}\label{sec:results}

We compare the performance of the three proposed conditioning methods with the two baselines and the state-of-the-art in Table~\ref{table:results}. 
We also provide results for the combination of our best method, attentional feature-based gating, and the second baseline model (\textit{emb. conc.}). 

The results show that incorporating external knowledge in RNN-based architectures consistently improves performance over the baseline for all datasets. 
Furthermore, feature-based gating improves upon baseline concatenation in the embedding layer across benchmarks, with the exception of \textit{PsychExp} dataset.

For the \textit{Sent17} dataset we achieve state-of-the-art $F_1$ score using the feature-based gating method; we further improve performance when combining gating with the \textit{emb. conc.} method. 
For \textit{SST-5}, we observe a significant performance boost with combined attentional gating and embedding conditioning (\textit{gate + emb. conc.}).
For \textit{PsychExp}, we marginally outperform the state-of-the-art also with the combined method, while for $Irony18$, feature-based gating yields the best results. 
Finally, concatenation based conditioning is the top method for $SCv1$, and the combination method for $SCv2$.  

\begin{figure*}[tbh]
        \begin{center}
        \begin{tikzpicture}
            \node at (0, 0) {\includegraphics[height=0.70cm]{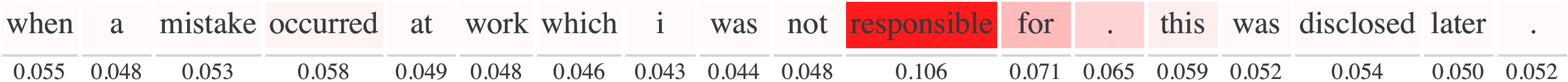}};
            \node at (0, -1) {\includegraphics[height=0.70cm]{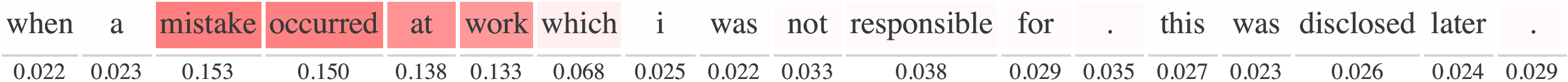}};

            \node at (7.5, 0) {\footnotesize{anger}};
            \node at (8.25, 0) {\includegraphics[height=0.3cm]{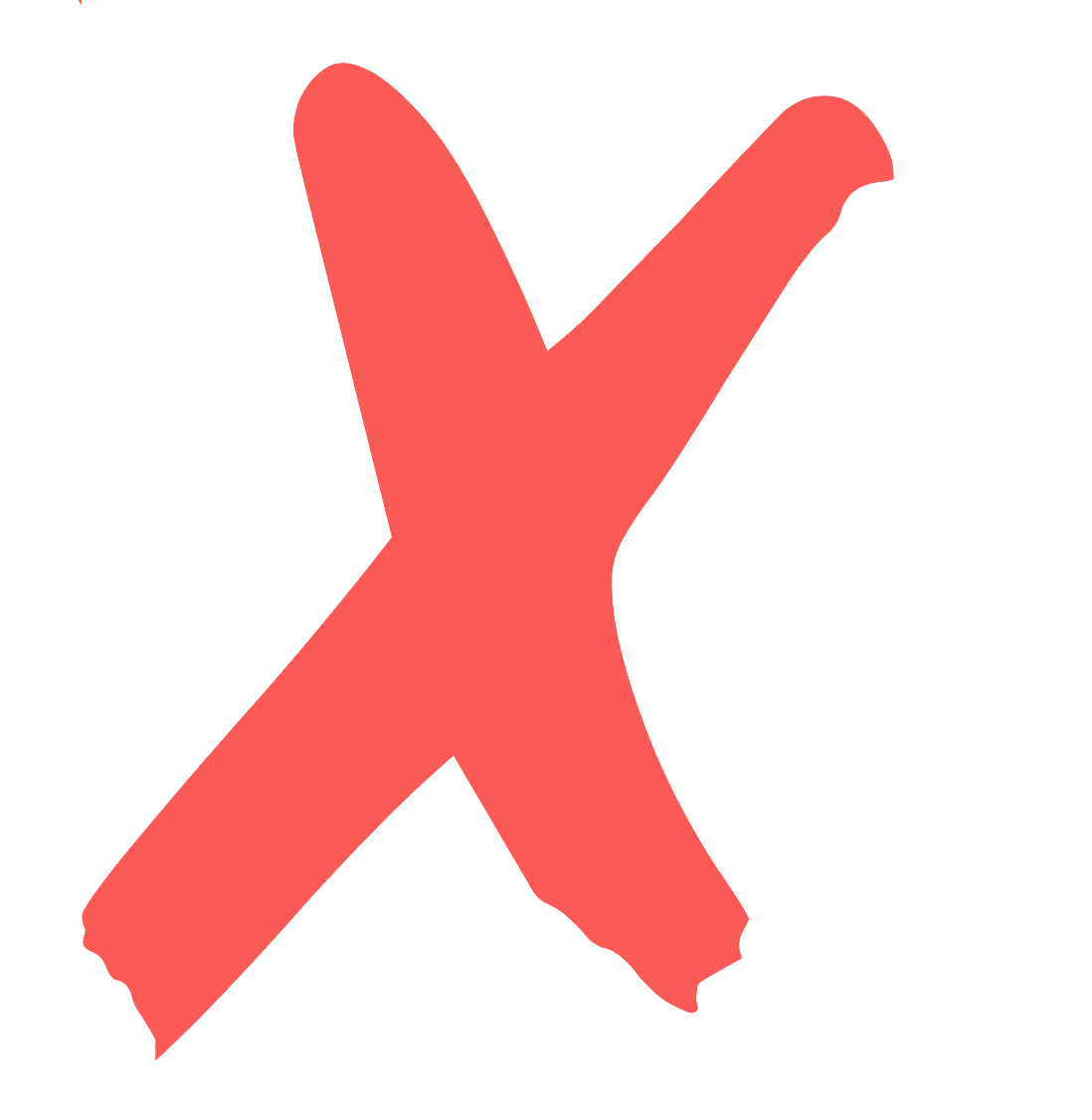}};
            \node at (7.5, -1) {\footnotesize{guilt}};
            \node at (8.25, -1) {\includegraphics[height=0.3cm]{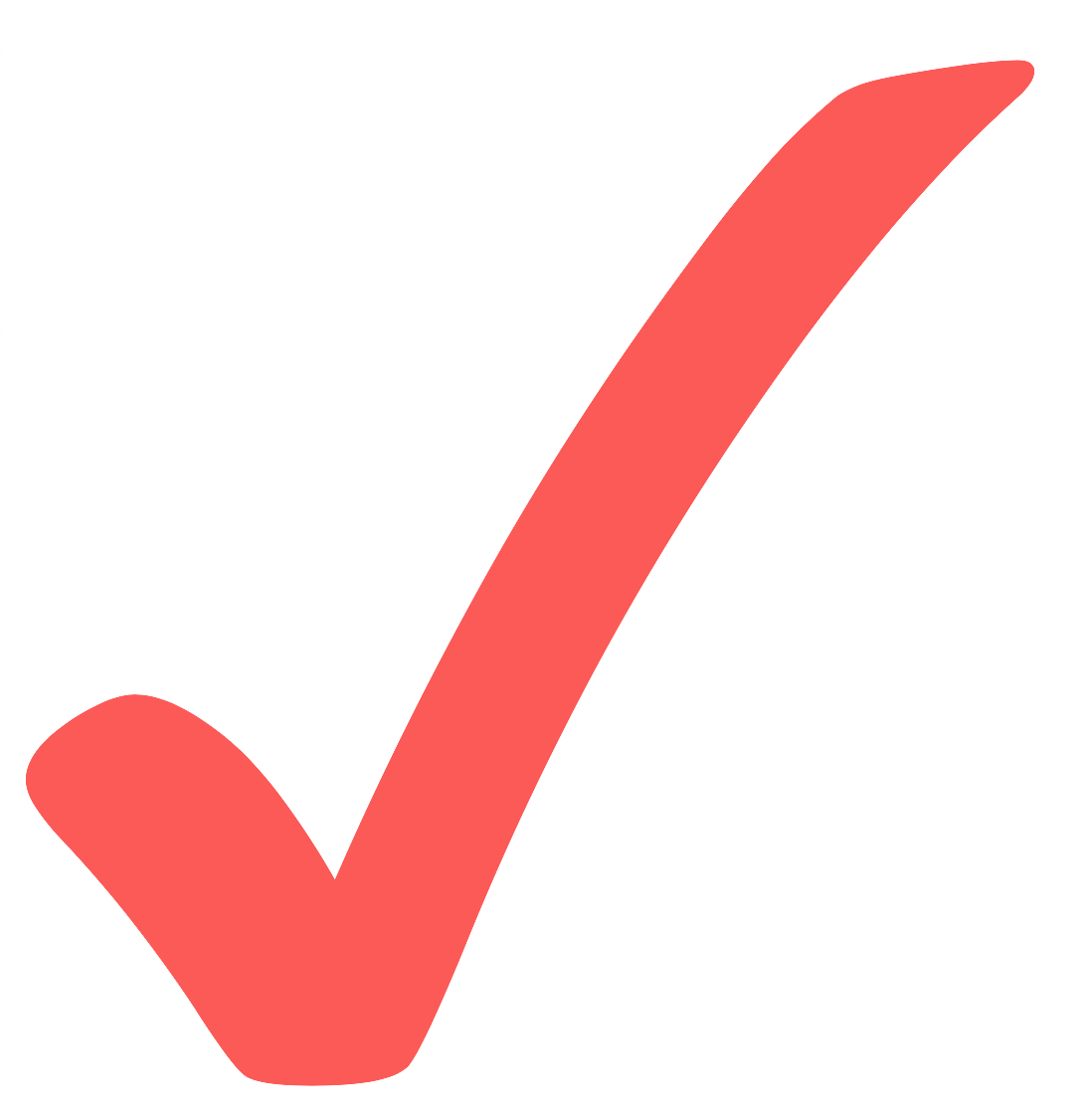}};

            \draw (-7.1, 0.6) -- (8.8, 0.6) -- (8.8, -1.6) -- (-7.1, -1.6) -- cycle;
        \end{tikzpicture}
        \end{center}
         \caption{Attention heatmap of a \textit{PsychExp} random test sample. The first attention distribution is created with the \textit{baseline} model without lexicon feature integration, while the second with the combination of our attentional feature-based gating method and the concatenation to word embeddings baseline (\textit{gate+emb.conc.}).}
    \label{fig:example}
\end{figure*}

Overall, attentional feature-based gating is the best performing conditioning method followed by concatenation. 
Attentional affine transformation underperforms, especially, for smaller datasets; this is probably due to the higher capacity of this model. 
This is particularly interesting since gating (Eq.~\ref{eq:f_gate}) is a special case of the affine transformation method (Eq.~\ref{eq:f_affine}), where the shifting vector $\beta$ is zero and the scaling vector $\gamma$ is bounded to the range $[0, 1]$ (Eq.~\ref{eq:gamma}). 
Interestingly, the combination of gating with traditional embedding-layer concatenation gives additional performance gains for most tasks, indicating that there are synergies to exploit in various conditioning methods. 

In addition to the performance improvements, we can visually evaluate the effect of conditioning the attention distribution on prior knowledge and improve the interpretability of our approach.
As we can see in Figure~\ref{fig:example}, lexicon features guide the model to attend to more salient words and thus predict the correct class.

\section{Conclusions \& Future work}
We introduce three novel attention-based conditioning methods and compare their effectiveness with traditional concatenation-based conditioning.
Our methods are simple, yet effective, achieving consistent performance improvement for all datasets.
Our approach can be applied to any RNN-based architecture as a extra module to further improve performance with minimal computational overhead.

In the future, we aim to incorporate more elaborate linguistic resources (e.g. knowledge bases) and to investigate the performance of our methods on more complex NLP tasks, such as named entity recognition and sequence labelling, 
where prior knowledge integration is an active area of research.

\section*{Acknowledgements}
We would like to thank our colleagues Alexandra Chronopoulou and Georgios Paraskevopoulos for their helpful suggestions and comments.
This  work  has  been  partially  supported  by  computational  timegranted  from  the  Greek  Research  \&  Technology  Network  (GR-NET) in the National HPC facility - ARIS. 
We thank NVIDIA for supporting this work by donating a TitanX GPU.




\bibliography{acl2019}
\bibliographystyle{acl_natbib}

\end{document}